\documentclass[journal]{IEEEtran}

\usepackage{amsmath,amsfonts}
\usepackage{algorithmic}
\usepackage{algorithm}
\usepackage{array}
\usepackage[caption=false,font=normalsize,labelfont=sf,textfont=sf]{subfig}
\usepackage{textcomp}
\usepackage{stfloats}
\usepackage{url}
\usepackage{verbatim}
\usepackage{graphicx}
\usepackage{soul}
\usepackage{xcolor}

\usepackage{multirow}
\usepackage{cite}
\usepackage{makecell, lipsum}
\usepackage{enumerate}
\usepackage{tabularx}
\usepackage{booktabs}
\usepackage{threeparttable}

\usepackage{graphicx}

\usepackage{textcomp}
\usepackage[table]{xcolor}
\usepackage[dvipsnames]{xcolor}
\usepackage{multicol}

\usepackage[caption=false,font=normalsize,labelfont=sf,textfont=sf]{subfig}
\usepackage{graphicx}

\usepackage{subcaption}
\let\oldnl\nl
\newcommand{\nonl}{\renewcommand{\nl}{\let\nl\oldnl}}
\usepackage{pifont}
\usepackage{array}
\newcolumntype{C}[1]{>{\centering\arraybackslash}m{#1}}
\usepackage{float}
\AtBeginDocument{%
  }

\usepackage[all=normal, paragraphs=tight, wordspacing=normal, charwidths=normal, indent=normal, floats=tight, leading=tight, lists=tight]{savetrees}

\begin{document}

\title{AgenticTwin: An Agentic LLM Framework Integrated with Digital Twin for Anomaly Detection}

\author{Touseef~Hasan,~\IEEEmembership{Student Member,~IEEE,}
        Mounika~Ghanta,~\IEEEmembership{Student Member,~IEEE,}
        Souvika~Sarkar,~\IEEEmembership{Member,~IEEE,}
        and~Ujjwal~Guin,~\IEEEmembership{Senior Member,~IEEE}
        
\thanks{Touseef Hasan and Souvika Sarkar are with the School of Computing, Wichita State University, KS, USA (e-mail: {txhasan4@shockers.wichita.edu and souvika.sarkar@wichita.edu}).}
\thanks{Mounika Ghanta and Ujjwal Guin are with the Department of Electrical and Computer Engineering, Auburn University, AL, USA (e-mail: {mzg0144 and ujjwal.guin}@auburn.edu).}
}

\maketitle

\begin{abstract}
Digital twins are increasingly used to monitor and simulate the behavior of cyber–physical systems. Even with skilled operators, interpreting anomalies detected within digital twin pipelines is challenging, as the sheer complexity and volume of raw sensor data make thorough analysis difficult. Recent advances in large language models (LLMs) offer promising capabilities for reasoning and explanation, yet their integration into digital twin-driven anomaly analysis remains underexplored. In this work, we propose \textit{AgenticTwin}, an agentic framework that integrates LLM-driven reasoning with a digital twin-based anomaly detection pipeline. The framework grounds LLM-generated explanations in outputs from a digital twin-driven anomaly classifier and enables human operators to ask relevant natural-language questions about the system. Beyond the framework itself, we introduce a benchmark-oriented evaluation pipeline constructed over synthetic anomalies injected into a real-world weather sensor dataset, enabling controlled generation of operator queries over anomaly events. We further evaluate the feasibility of deploying lightweight, open-source LLMs for practical cyber-physical environments. Experimental results demonstrate that structured agent collaboration and knowledge-grounded reasoning improve diagnosis quality, contextual retrieval, and mitigation quality across diverse possible anomaly scenarios.

\end{abstract}


\section{Introduction}
\label{section: intro}

\begin{figure} [t] \centerline{\includegraphics[width=1\linewidth, trim=500 300 540 50, clip]{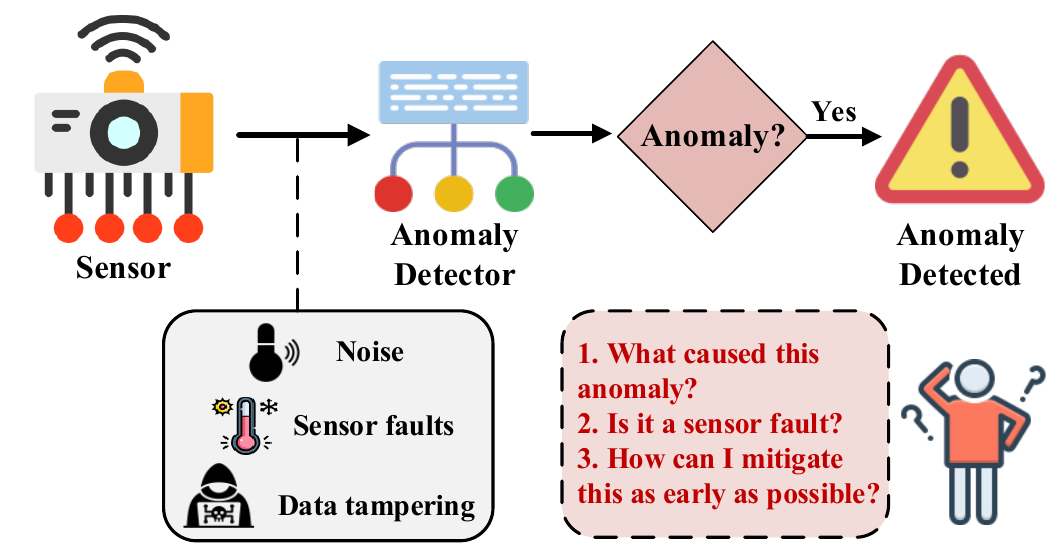}}
    \caption{Traditional anomaly detection without actionable explanations or mitigation guidance.}
    \label{fig:teaser}
\end{figure}

\IEEEPARstart{D}{igital} twins (DTs) are virtual replicas of physical systems for monitoring, modeling, and managing cyber–physical systems (CPS) across domains like manufacturing, healthcare, and industrial automation~\cite{grieves2016digital, tao2019digital, lu2020digital, cimino2019review, vallee2023digital}. DTs aid operators by continuously observing operational states, simulating system behavior, and proactively identifying anomalies~\cite{airehenbuwa2025advancing}. However, communicating these anomalies to users remains a significant challenge as such systems typically lack interpretability~\cite{sarker2024explainable}. Figure~\ref{fig:teaser} presents the limitations of traditional anomaly detection within DT architectures. Although it is possible to accurately flag anomalous events with such frameworks, they provide limited interpretability regarding: ($i$) why an anomaly occurred, ($ii$) what underlying mechanisms may have contributed to it, and/or ($iii$) what corrective actions should be taken to mitigate it~\cite{huang2021digital, kobayashi2024explainable}. As a result, skilled human operators are frequently required to manually analyze raw sensor readings and system logs to diagnose faults and determine mitigation strategies. This process is becoming increasingly difficult as CPS environments become more complex and data-intensive~\cite{xu2023digital, eckhart2019digital} with time.

The enhanced reasoning capabilities of large language models (LLMs) have inspired promising applications in DTs, transforming raw numbers into interpretable explanations~\cite{yang2025leveraging, lekshmi2026integrating, ahmadpanah2026semantic, wu2026large, khaldi2026agentic}. However, deploying LLMs directly within CPS workflows presents several challenges. First, LLMs suffer from hallucinations when they lack access to reliable domain-specific knowledge or context describing the current system state~\cite{greis2025multi, pan2026llm}. Second, monolithic LLM-based reasoning frameworks often struggle to jointly perform anomaly diagnosis, mitigation planning, and user interaction within a single prompt, resulting in inconsistent or incomplete anomaly analysis~\cite{fu2026generative}. Finally, many existing approaches rely on large proprietary models with substantial computational requirements, limiting practical deployment in resource-constrained CPS~\cite{hong2024llm, mata2026digital}.

To address these challenges, we propose \textit{AgenticTwin}, an agentic LLM framework integrated with a DT–driven anomaly detection pipeline. Instead of relying on a single monolithic LLM, our framework decomposes anomaly reasoning into task-specific agents responsible for diagnosis, mitigation planning, and contextual historical retrieval. These agents operate on structured evidence generated by the DT and the anomaly classifier, enabling interpretable reasoning about anomalous system behavior. To systematically evaluate this agentic framework, we also construct a benchmark-oriented anomaly analysis dataset by injecting synthetic anomalies into a real-world weather sensor dataset. The benchmark enables controlled and reproducible evaluation of anomaly diagnosis, historical retrieval, and mitigation reasoning across diverse LLM architectures and deployment settings. The following research questions drive the formulation and design of the proposed framework: ($i$) whether LLM-based reasoning can be grounded in context through a DT-enabled anomaly classifier to support interpretable and reliable anomaly diagnosis in CPS; ($ii$) whether decomposing anomaly reasoning into specialized agents improves diagnosis, retrieval, and mitigation quality compared to a monolithic LLM framework; and ($iii$) whether lightweight, open-source LLMs can enable scalable and practical deployment of agentic reasoning frameworks in resource-constrained cyber-physical environments. Our key contributions are as follows:

\begin{itemize}
    \item We propose \textit{AgenticTwin}, an agentic LLM framework integrated with a DT-driven anomaly detection pipeline for interpretable anomaly analysis in cyber-physical systems. The framework combines DT predictions, anomaly diagnosis, historical retrieval, and mitigation reasoning within a coordinated agentic architecture.
    \item We develop a benchmark-oriented evaluation dataset by injecting synthetic anomalies into a real-world weather sensor dataset. The benchmark consists of synthetic operator queries generated over anomalous events and enables controlled evaluation of diagnosis quality, retrieval capability, and mitigation reasoning.
    \item We conduct a comparative assessment between agentic and monolithic LLM reasoning architectures to investigate whether task decomposition improves anomaly interpretation, contextual retrieval, and mitigation quality in DT-enabled cyber-physical environments. Additionally, we evaluate the practicality of lightweight, open-source LLMs for deployment in resource-constrained settings.
\end{itemize}

The rest of the paper is organized as follows: Section~\ref{section: background} covers prior work on anomaly detection, LLM-integrated DT frameworks, and agentic architectures. Section~\ref{section: framework} presents the proposed framework. Section~\ref{section: experiment} contains our experimental findings and Section~\ref{section: conclusion} concludes the paper.

\section{Prior Work}
\label{section: background}

This section begins with a brief description of digital twins, followed by an examination of anomaly detection in the CPS environment and frameworks that integrate DTs with LLMs.

\subsection{Digital Twins and Anomaly Detection in CPS}

A digital twin is a synchronized digital representation of a physical device, process, or service that continuously updates its current state using real-time observations collected from physical sensors~\cite{qi2021enabling, semeraro2021digital}. DTs have been widely adopted in CPS for monitoring and analysis~\cite{gill2025leveraging, haghshenas2023predictive}. In practice, DTs are commonly constructed using a combination of data-driven machine learning models and domain-specific physics-based rules~\cite{ritto2021digital}. Machine learning models capture complex relationships from historical sensor observations, while physics-guided constraints enforce consistency with known system behavior and operational limits~\cite{chen2023advance, aivaliotis2019methodology}. The combination of these two components enables DTs to estimate expected system states and identify deviations between observed and predicted behavior, which can subsequently be used for anomaly detection and diagnosis~\cite{moharam2025anomaly}. For anomaly detection, several studies have explored various machine learning algorithms by identifying deviations from the learned patterns~\cite{zhao2024sgad, qiu2025tab}. While these models identify and classify anomalous patterns, they may not fully capture the behavior of the underlying physical system. Most of these approaches are heavily data-driven and completely rely on sensor measurements, which may reduce reliability under noisy or corrupted sensor conditions~\cite{rahim2024digital}. Existing research has also proposed several techniques, such as threshold-based filtering and rule-based fault handling~\cite{ma2024digital},~\cite{ahmed2025weather}. Although these methods help reduce the impact of faulty sensor data by fault identification and classification, they provide limited support for interactive user-oriented contextual anomaly reasoning, diagnosis, and mitigation assistance in CPS environments.

\subsection{Integration of LLMs in Digital Twin Frameworks}

Traditionally, DTs have relied on structured inputs and outputs, requiring domain-specific knowledge to interpret and interact with the simulated environment~\cite{xia2025architecture}. However, the emergence of LLMs has opened new avenues for embedding reasoning, explanation, and decision-support capabilities directly into DT frameworks~\cite{huang2024simulation, ferdousi2026defecttwin, li2026lsdts, kampourakis2026systematic}. Recent studies have explored the potential of LLM-driven DTs in network optimization, agriculture, industrial automation, healthcare, and smart cities~\cite{guo2025survey, zhang2024large, xia2024fcllm, gautam2025iiot, makarov2025large, vahdati2025multi, mandal2024llmasmmkg, xu2026towards, nechesov2025virtual, deng2026integrating}. While most of these frameworks use LLMs to primarily interpret and explain DT outputs to end users, some use LLMs to simulate complex environments and therefore generate the DT itself~\cite{yang2024llm, wang2026simbench}. While promising, these efforts remain limited to static or post hoc analysis without real-time user interaction or closed-loop reasoning with the DT~\cite{xu2023generative}. Moreover, these approaches mostly rely on large-scale or proprietary LLMs, which require significant computational resources, limiting their feasibility in resource-constrained cyber-physical environments.

\subsection{Agentic Architectures for Autonomous Decision Making in Digital Twins}

As LLMs become increasingly integrated into cyber-physical and DT systems, agentic architectures have emerged as a promising approach for decomposing complex decision-making tasks into specialized and collaborative reasoning components~\cite{pan2026llm}. Agentic systems in safety-critical environments, such as manufacturing, emphasize effective agent collaboration to ensure safe and reliable operation~\cite{gill2025leveraging, sun2024empowering, li2025future, ling2026sr}. Prior studies have stated the importance of a user interface that allows users to interact with the DT~\cite{kalyani2024role}. The agents must not only coordinate but also validate, explain, and justify their actions, especially when recommendations affect physical assets or operational safety~\cite{nicoletti2025digital, hasan2026integrating}. Agentic LLMs have been explored to automate the parametrization of simulation models in DTs, iteratively control temperature on microcontrollers, and autonomously select optimal tools and data streams for user-specific queries~\cite{xia2024llm, vyas2025autonomous, timms2025agentic}. However, most agentic frameworks fail to validate their generated responses, which reinforces concerns about hallucination due to inadequate context or a lack of domain-specific evidence.

\section{The Proposed AgenticTwin Framework}
\label{section: framework}

The limitations of traditional anomaly detection systems discussed in Section~\ref{section: intro} motivate the need for a framework that not only detects anomalous behavior but also explains its causes, retrieves relevant contextual evidence, and recommends corrective actions. To address these challenges, we propose \textit{AgenticTwin}, a DT-integrated agentic LLM framework for interpretable anomaly analysis in CPS. The framework combines DT-based anomaly detection with a curated anomaly knowledge base, an anomaly repository, and a coordinated multi-agent reasoning interface. Together, these components enable anomaly diagnosis, contextual retrieval of previously observed events, and mitigation planning while supporting deployment with lightweight, open-source LLMs.

\begin{figure*}[t]
\centering
\centerline{\includegraphics[width=1\linewidth, trim = 20 20 20 60]{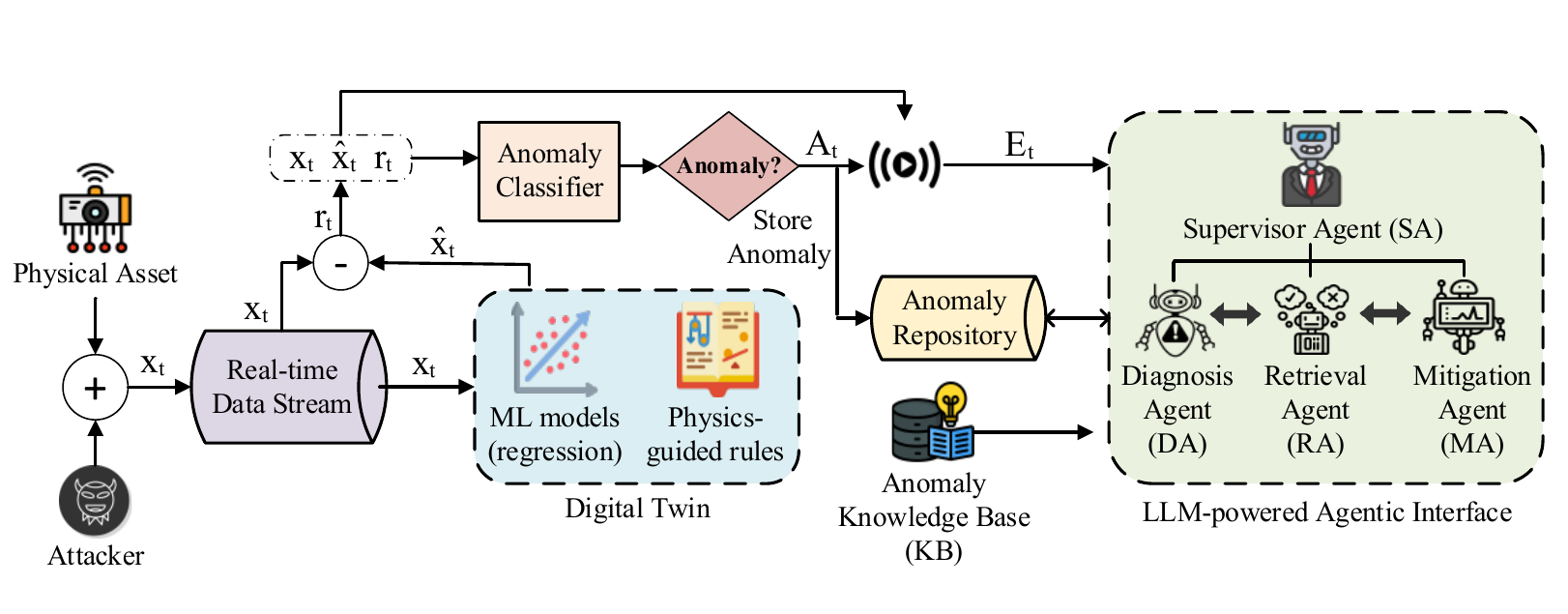}}
 \caption{Overview of the proposed \textit{AgenticTwin} architecture, integrating a DT with an LLM-powered multi-agent interface for interpretable anomaly analysis in CPS.}
\label{fig:framework}
\end{figure*}
 
Figure~\ref{fig:framework} provides an overview of the proposed architecture. The overall workflow begins with a real-time data stream collected from the physical asset. In the operational setting, the incoming sensor vector \(\mathbf{x}_t\) is directly passed to both the DT and the anomaly classifier. The DT estimates the expected system state \(\hat{\mathbf{x}}_t\) using machine learning regression models and physics-guided rules. The residual vector \(\mathbf{r}_t\) is then computed and used by the anomaly classifier to determine whether the current observation is anomalous. The attacker block in Figure~\ref{fig:framework} represents the benchmark construction setting, where anomalous behavior is introduced into historical sensor data to simulate corrupted or tampered measurements. This allows the framework to be trained and evaluated under controlled anomaly scenarios. During online execution, however, the system does not assume prior knowledge of whether the incoming data has been tampered with; it only observes \(\mathbf{x}_t\), estimates \(\hat{\mathbf{x}}_t\), computes \(\mathbf{r}_t\), and relies on the anomaly classifier to infer the anomaly output \(A_t\). Once an anomaly is detected, the classifier produces \(A_t\), i.e., the predicted anomaly state and label. The system then forms a structured anomaly event \(E_t\), consisting of the timestamp, observed values, predicted values, residuals, and predicted anomaly label. This event is stored in the anomaly repository and passed to the LLM-powered agentic interface. The Retrieval Agent (RA), Diagnosis Agent (DA), Mitigation Agent (MA), and Supervisor Agent (SA) then use \(E_t\), the predicted data \(\hat{\mathbf{x}}_t\), the anomaly repository, and the anomaly knowledge base (KB) to generate contextual diagnosis and mitigation support for the operator. Unlike conventional monolithic LLM frameworks, the \textit{AgenticTwin} framework delegates the following specialized LLM agents with distinct responsibilities:

\begin{itemize}
    \item \textit{$DA$ -- Diagnosis Agent:} This agent provides reasoning of the underlying cause of the detected anomalies. Most anomaly detection methods focus on indicating the presence of abnormal behavior in the data but provide limited insights into the reasons behind it. Detection of an anomaly alone is insufficient for effective decision-making, as the operator needs to understand the underlying factors to implement corrective measures. To address this limitation, the DA uses DT predictions, residual patterns, and anomaly taxonomy information to identify potential root causes.
    
    \item \textit{$RA$ -- Retrieval Agent:} This agent retrieves and ranks the previous similar anomaly cases from a historical anomaly repository. The rationale for the RA is to provide additional context by leveraging recurring and past anomaly patterns, thereby improving understanding of current system behavior and supporting accurate diagnosis and mitigation decisions.
 
    \item \textit{$MA$ -- Mitigation Agent:} This agent recommends corrective actions to address the detected anomaly and minimize its impact on the system. Based on the diagnosis and retrieved historical evidence, the MA suggests the necessary actions.
    
    \item \textit{$SA$ -- Supervisor Agent:} This agent is responsible for coordinating the activities of the specialized agents and ensuring that their outputs are combined into a coherent response. The $SA$ collects and reviews the outputs generated by other agents and integrates them into a unified explanation. It serves as the final coordination layer.
\end{itemize}

The proposed framework is designed as a modular and generalizable architecture that can operate with lightweight, open-source LLMs, enabling deployment in resource-constrained cyber-physical environments. The following subsections describe each of the major components of the framework.

\subsection{Digital Twin}
\label{subsection:dt}

The digital twin (DT) serves as a physics-informed virtual representation of a physical entity. It comprises a set of domain-specific physical laws and constraints, along with the structural relations among the variables of a system. A cyber-physical system (CPS) equipped with \(n\) sensors can be represented by an observed state vector at time \(t\)~\cite{course2023state}. This state-vector representation provides a compact description of the system condition using all sensor measurements available at a given timestamp. Formally,

\begin{equation} 
\mathbf{x}_t = [x_{1,t},x_{2,t}, \ldots,x_{n,t}]^{T} 
\in \mathbb{R}^{n}, 
\label{eq:general_state} 
\end{equation} 

where \(x_{i,t}\) denotes the reading of the \(i^{th}\) sensor at time \(t\). In the proposed framework, we instantiate this formulation using a weather monitoring CPS consisting of three environmental variables: temperature, dew point temperature, and relative humidity~\cite{beutenberg_data}. Therefore, the observed state vector becomes:

\begin{equation} 
\mathbf{x}_t = [T_t,T_{d,t},RH_t]^{T} 
\in \mathbb{R}^{3}, 
\label{eq:weather_state} 
\end{equation} 

where \(T_t\), \(T_{d,t}\), and \(RH_t\) denote temperature, dew point temperature, and relative humidity at time \(t\), respectively.

The objective of the DT is to estimate the expected system state \(\hat{\mathbf{x}}_t\) from previously observed system behavior. Since environmental variables often exhibit temporal dependencies as well as diurnal and seasonal patterns, the DT prediction model uses both past observations and a temporal encoding function~\cite{taylor2018forecasting, lim2021temporal}. The expected state is modeled as follows:

\begin{equation} 
\hat{\mathbf{x}}_t = 
\mathcal{F}_{\theta} 
\left( 
\mathbf{x}_{1:t-1},\phi(t) 
\right), 
\label{eq:dt_prediction} 
\end{equation}

where \(\mathbf{x}_{1:t-1}\) denotes the historical sensor sequence before time \(t\), \(\phi(t)\) denotes the temporal encoding function capturing periodic characteristics such as time-of-day and seasonal effects, and \(\mathcal{F}_{\theta}(\cdot)\) represents the regression-based DT model parameterized by \(\theta\)~\cite{zia2025physics, prantikos2022physics}. In this work, the DT follows a hybrid modeling strategy in which data-driven regression captures statistical dependencies among sensor variables, while physics-guided constraints incorporate known relations among environmental variables. The model parameters are learned offline from historical training data by minimizing the difference between the observed state and the DT-estimated state. This training objective is written as:

\begin{equation} 
\theta^{*} = \arg\min_{\theta} \mathcal{L}_{DT}(\theta),
\label{eq:dt_argmin} 
\end{equation} 

where the DT prediction loss is defined as:

\begin{equation}
\mathcal{L}_{DT}(\theta) = \frac{1}{N} \sum_{t=1}^{N} 
\left\| 
\mathbf{x}_t - \hat{\mathbf{x}}_t 
\right\|_2^2,
\label{eq:dt_loss} 
\end{equation} 

where \(N\) denotes the number of training samples, \(\|\cdot\|_2\) denotes the Euclidean norm, and \(\mathcal{L}_{DT}\) measures the average prediction error between the observed and estimated system states. Thus, Eq. (\ref{eq:dt_argmin}) indicates that the optimal DT parameters \(\theta^{*}\) are obtained by minimizing the loss defined in Eq. (\ref{eq:dt_loss}). After training, the DT provides \(\hat{\mathbf{x}}_t\), which serves as the expected system state used by the anomaly classifier to identify deviations from normal behavior.

\subsection{Anomaly Classifier}
The objective of the anomaly classifier is to distinguish normal and abnormal behavior of the system by analyzing the discriminating patterns in the data. Beyond binary detection, this module further identifies the fault type based on the relationships between variables in the system~\cite{hsu2023comparison, zhang2019deep}. The proposed anomaly classifier infers faults by jointly considering observed system values, DT estimates, and the corresponding residual information. The DT-informed input can be defined as:
\begin{equation}
    \mathbf{z}_t=
    \left[
    \mathbf{x}_t,
    \hat{\mathbf{x}}_t,
    \mathbf{r}_t
    \right],
    \label{eq:feature}
\end{equation}

where $r_t$ is the residual vector that captures discrepancies between observed and expected system behavior:

\begin{equation}
\mathbf{r}_t =
\mathbf{x}_t-\hat{\mathbf{x}}_t.
\label{eq:residual}
\end{equation} 

The anomaly classification task can be formulated as:

\begin{equation}
A_t = (a_t, \hat{y}_{t}), 
\label{eq:classifier}
\end{equation}

where $a_t \in \{0,1\}$ denotes the anomaly state and $\hat{y}_{t}$ represents the anomaly label~\cite{jan2021distributed}. Since multiple sensor faults may occur simultaneously, $\hat{y}_t$ may contain more than one anomaly label in multi-fault scenarios. The classifier is trained to accurately predict the anomaly labels by minimizing:

\begin{equation}
\mathcal{L}_A=l(y_t,\hat{y}_t),
\end{equation}
\begin{equation}
    \psi^{*}=\arg\min_{\psi}\mathcal{L}_{A},
\end{equation}

where $y_t$ and $\hat{y}_t$ denote ground truth and predicted anomaly labels respectively and $l(\cdot)$ denotes the model loss function~\cite{saeed2021cafd}. When an anomaly is detected (\(a_t=1)\), the framework generates a structured anomaly event:
\begin{equation}
E_t =
\left(
\tau_t,
\mathbf{x}_t,
\hat{\mathbf{x}}_t,
\mathbf{r}_t,
A_t
\right),
\label{eq:event}
\end{equation}

where \(\tau_t\) denotes the event timestamp. The anomaly event \(E_t\) serves as the primary information unit exchanged within \textit{AgenticTwin} and is subsequently stored in the historical anomaly repository and passed to the LLM-powered agentic interface for retrieval, diagnosis, and mitigation reasoning.

\subsection{Anomaly Knowledge Base}
\label{subsection:kb}

To support grounded anomaly reasoning, we construct a curated anomaly knowledge base (KB) that contains structured descriptions of anomaly types observed in the benchmark environment. Existing CPS knowledge bases, such as {MITRE ATT\&CK for ICS~\cite{mitre}} primarily focus on adversarial attack behaviors and industrial threat modeling, but do not directly provide descriptions tailored to sensor-level time-series anomalies. Therefore, we develop a lightweight anomaly taxonomy specifically designed for DT-driven anomaly analysis tasks. This knowledge base is designed as a modular and extensible component that can be replaced or expanded for different CPS environments. Rather than relying on a fixed anomaly ontology, the framework allows users to incorporate domain-specific anomaly taxonomies in a plug-and-play manner. The current taxonomy contains the following four anomaly types corresponding to the injected anomaly scenarios used in the benchmark dataset, adapted from~\cite{jan2021distributed}:

(i) Spike: Defined as a sudden jump or drop in the signal affecting one or a few consecutive samples while the surrounding measurements remain within their normal range:
\begin{equation}
x_{\mathrm{faulty}}(t) = x(t) + n(t),
\label{eq:spike_fault}
\end{equation}
where \(n(t)\) represents the noise applied to the sensor data at time \(t\).

(ii) Drift: A gradual deviation from the normal range over time. Drift faults induce a slow bias that results in a sustained offset between the measured and the true values:
\begin{equation}
x_{\mathrm{faulty}}(t) = x(t) + a \cdot t,
\label{eq:drift_fault}
\end{equation}
where \(a\) controls the rate of change, leading to a gradual deviation from the original values.

(iii) Stuck-at: Occurs when the sensor output becomes frozen for consecutive intervals:
\begin{equation}
x_{\mathrm{faulty}}(t) = c, \quad \forall t \in [t_1,t_2],
\label{eq:stuck_fault}
\end{equation}
where \(c\) is the constant value assigned to the faulty sensor reading during the interval \([t_1,t_2]\).

(iv) Replay: Occurs when current sensor values are replaced with previously recorded data over a time interval:
\begin{equation}
x_{\mathrm{faulty}}(t) = x(t - \Delta t),
\label{eq:replay_fault}
\end{equation}
where \(\Delta t\) represents the time lag.

\begin{figure} [t]
\centerline{\includegraphics[width=1\linewidth]{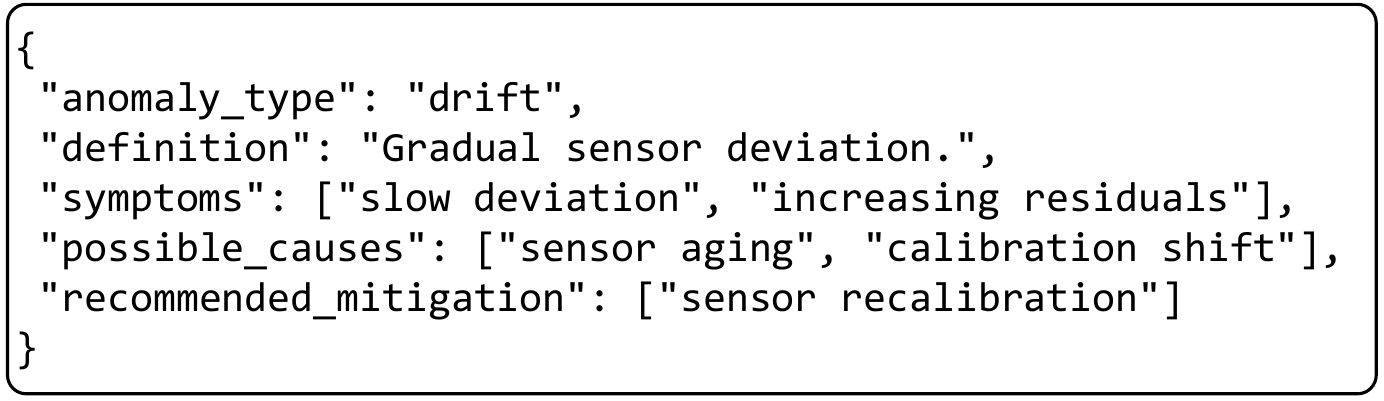}}
\caption{Entry in the anomaly knowledge base for a drift fault.}
\label{fig:examplekb}
\end{figure}

Each anomaly entry contains structured semantic information, including the anomaly definition, observable symptoms, possible causes, and recommended mitigation actions. The knowledge base is stored in a structured JSON format and queried during inference to provide retrieval-augmented grounding for the diagnosis and mitigation agents. To improve reliability, the knowledge base was manually reviewed and refined using domain knowledge and existing literature on anomaly analysis. An example knowledge base entry is shown in Figure~\ref{fig:examplekb}, specifying the anomaly type, its definition, observable symptoms, possible causes, and recommended mitigation actions in a structured JSON format.

\begin{figure} [ht] \centerline{\includegraphics[width=1\linewidth]{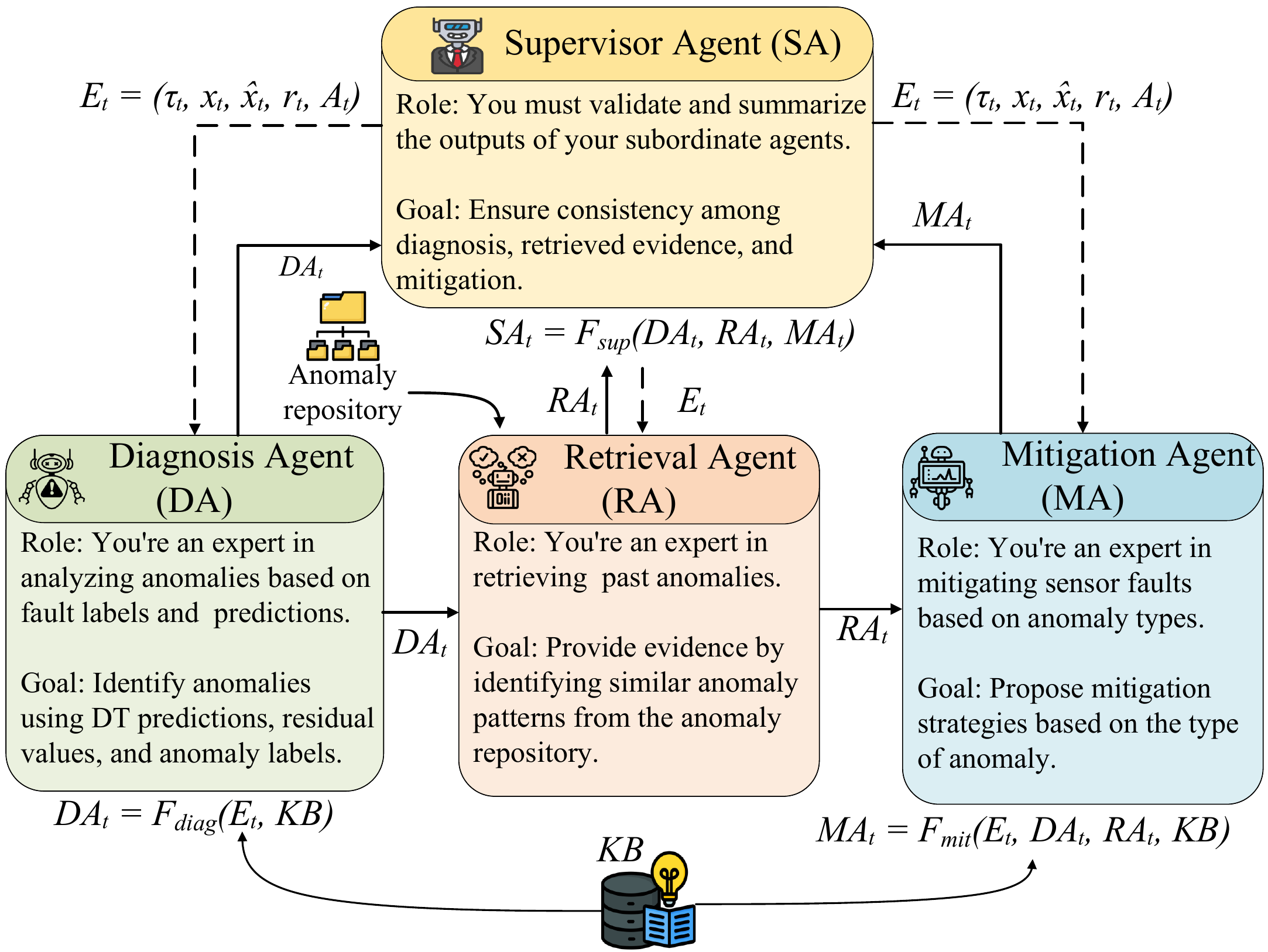}}
\caption{Modular breakdown of the predefined roles, goals, inputs, and output flows of the agents in \textit{AgenticTwin}.}
\label{fig:agents}
\end{figure}

\subsection{LLM-Powered Agentic Interface}

At the core of the \textit{AgenticTwin} framework is the multi-agent LLM interface, forming the reasoning layer of the proposed framework. Its primary objective is to transform structured anomaly events into interpretable diagnosis explanations, contextual historical evidence, and mitigation recommendations for human operators. Rather than relying on a single monolithic LLM, \textit{AgenticTwin} decomposes anomaly reasoning into multiple specialized agents coordinated through a supervisory architecture. Figure~\ref{fig:agents} illustrates the four-stage agentic reasoning workflow used in \textit{AgenticTwin}: diagnosis, context retrieval, mitigation, and supervision.

The DA receives the structured anomaly event, \(E_t\) and the anomaly knowledge base, KB. Its objective is to identify the anomaly type and generate grounded root-cause explanations. The output of the DA is represented as:

\begin{equation}
DA_t = \mathcal{F}_{\mathrm{diag}}(E_t,KB).
\label{eq:diagnosis_agent}
\end{equation}

The RA retrieves and ranks previously observed anomaly cases that match the predicted anomaly type of the current event. Given an anomaly event \(E_t\), the RA filters the historical anomaly repository using the predicted anomaly label(s) and returns only the records that share the same anomaly type. For single-label anomaly events, this reduces to retrieving historical cases with the same predicted anomaly type. For multi-label events, the agent retrieves previous cases that share at least one anomaly label with the current event. The RA then ranks these records according to their relevance to \(E_t\) by considering the observed sensor values, DT predictions, and residual patterns. The final ranked set is denoted by \(RA_t\).

The MA generates corrective actions based on the DA's outputs, retrieved historical evidence from the RA, and anomaly taxonomy information from the curated anomaly knowledge base (KB). The output of the MA is represented as:
\begin{equation}
MA_t =
\mathcal{F}_{\mathrm{mit}}
(E_t, DA_t, RA_t, KB).
\label{eq:mitigation_agent}
\end{equation}

Finally, the SA coordinates the overall reasoning workflow of the agentic interface and synthesizes the final explanation as the response to the user. The output of the SA is represented as:
\begin{equation}
SA_t =
\mathcal{F}_{\mathrm{sup}}
(DA_t, RA_t, MA_t).
\label{eq:supervisor_agent}
\end{equation}

\subsection{Evaluation Metrics}

Different evaluation metrics are used to assess the various reasoning tasks performed by the LLM agents in \textit{AgenticTwin}.

\subsubsection{Diagnosis Agent and Mitigation Agent Evaluation} The DA and MA are evaluated by calculating the semantic similarity between the agents' generated responses and the reference responses in our benchmark. Given a generated response \(g_i\) and a reference response \(r_i\), their semantic similarity is computed using cosine similarity:

\begin{equation}
\mathrm{Sim}(g_i,r_i)
=
\frac{
f(g_i)\cdot f(r_i)
}{
\|f(g_i)\| \|f(r_i)\|
},
\label{eq:semantic_similarity}
\end{equation}

where \(f(\cdot)\) denotes the sentence embedding function.

\subsubsection{Historical Retrieval Agent Evaluation} The RA is evaluated using Precision@\(k\), Recall@\(k\), and Mean Reciprocal Rank (MRR) as follows:


\begin{equation}
Precision@k = \frac{|R_k \cap \hat{R}_k|}{k},
\label{eq:precision_k}
\end{equation}


\begin{equation}
Recall@k = \frac{|R_k \cap \hat{R}_k|}{|R|},
\label{eq:recall_k}
\end{equation}


\begin{equation}
MRR = \frac{1}{Q}\sum_{i=1}^{Q}\frac{1}{r_i},
\label{eq:mrr}
\end{equation}

where \(R_k\) denotes the set of relevant historical anomaly records, \(\hat{R}_k\) denotes the set of top-\(k\) retrieved records, \(R\) denotes the complete set of relevant records, \(Q\) is the number of retrieval queries, and \(r_i\) is the rank position of the first relevant record retrieved for query \(i\).

\section{Experimental Results and Discussion}
\label{section: experiment}

This section presents the experimental methodology and results used to evaluate the proposed \textit{AgenticTwin} framework.

\subsection{CPS Testbed and Benchmark Construction}

\begin{table}
\centering
\caption{Overview of the experimental dataset.}
\label{tab:dataset}
\fontsize{8.5}{9.5}\selectfont
\setlength{\tabcolsep}{2.25pt}
\renewcommand{\arraystretch}{1}
\begin{tabular}{@{}lcccc@{}}
\toprule
\textbf{Feature} &
\textbf{Total samples} &
\textbf{Training samples} &
\textbf{Testing samples} &
\textbf{Variables} \\
\midrule
Count
& 34,991
& 27,992
& 6,999
& 3 \\
\bottomrule
\end{tabular}
\end{table}

We evaluate \textit{AgenticTwin} using a publicly available weather dataset~\cite{kolle2008documentation, ahmed2025weather}. We use meteorological observations collected by the Beutenberg weather station operated by the Max Planck Institute for Biogeochemistry in Jena, Germany~\cite{beutenberg_data}. The processed dataset contains a total of 34,991 samples collected at 10-minute intervals between January 1 and August 31, 2025. The dataset is split using an 80\%-20\% ratio into training and test sets, having 27,992 and 6,999 samples respectively. From the dataset, we consider three variables for our experiment, i.e., temperature, dew point temperature, and relative humidity. An overview of the dataset is outlined in Table~\ref{tab:dataset}.

Synthetic anomalies are injected into the training set to train the anomaly classifier and a separately injected held-out testing set is used for evaluation. Following the formulations introduced in Section~\ref{subsection:kb}, we generate four anomaly categories: spike, drift, stuck-at, and replay. The anomalies are introduced across different sensors and timestamps to represent both transient and persistent faults. We construct single-fault scenarios, in which at most one anomaly occurs at a timestamp, and multi-fault scenarios, in which multiple anomaly types may simultaneously affect different sensor variables. The resulting benchmark contains normal and anomalous observations together with structured metadata, including timestamps, observed and DT-predicted values, residuals, and ground-truth anomaly labels.

\subsection{Anomaly Classification Models and Performance}

The anomaly classifier is responsible for identifying abnormal sensor behavior using DT-grounded evidence. Instead of relying solely on raw sensor values, the classifier operates on observed sensor measurements, DT-predicted values, and residual values described in Section \ref{subsection:dt}. The primary purpose of the anomaly classifier within the \textit{AgenticTwin} framework is the generation of structured evidence that is used by the LLM-based reasoning layer. The classifier outputs the predicted anomaly state and labels. We evaluate multiple classification models, including Support Vector Machines (SVM)~\cite{cortes1995support}, Random Forest (RF)~\cite{ho1995random}, eXtreme Gradient Boosting (XGBoost)~\cite{chen2016xgboost}, Light Gradient Boosting Machines (LightGBM)~\cite{ke2017lightgbm}, Multi-Layer Perceptron (MLP)~\cite{murtagh1991multilayer}, Time Series Transformer (TST)~\cite{li2019enhancing}, PatchTST~\cite{nie2022time}, Timer-S1~\cite{liu2026timer}, and TimesNet~\cite{wu2022timesnet}. We evaluated the models using observed sensor values, DT predictions, and residuals. We further evaluated using both the single-fault (SF) and multi-fault (MF) scenarios.

\begin{table*}[t]
\centering
\footnotesize
\setlength{\tabcolsep}{5pt}
\caption{Performance of different classification models under single-fault (SF) and multi-fault (MF) anomalous conditions.}
\renewcommand{\arraystretch}{1.25}
\begin{tabular}{l|cc|cc|cc|cc|cc|cc|cc|cc|cc}
\hline
\multirow{2}{*}{\textbf{Metric}}
& \multicolumn{2}{c|}{\textbf{MLP}}
& \multicolumn{2}{c|}{\textbf{SVM}}
& \multicolumn{2}{c|}{\textbf{RF}}
& \multicolumn{2}{c|}{\textbf{XGBoost}}
& \multicolumn{2}{c|}{\textbf{LightGBM}}
& \multicolumn{2}{c|}{\textbf{TST}}
& \multicolumn{2}{c|}{\textbf{PatchTST}}
& \multicolumn{2}{c|}{\textbf{Timer-S1}}
& \multicolumn{2}{c}{\textbf{TimesNet}} \\
\cline{2-19}
& \textbf{SF} & \textbf{MF}
& \textbf{SF} & \textbf{MF}
& \textbf{SF} & \textbf{MF}
& \textbf{SF} & \textbf{MF}
& \textbf{SF} & \textbf{MF}
& \textbf{SF} & \textbf{MF}
& \textbf{SF} & \textbf{MF}
& \textbf{SF} & \textbf{MF}
& \textbf{SF} & \textbf{MF} \\
\hline
Accuracy  
& \textit{0.98} & \textit{0.94} & 0.70 & 0.71 & 0.89 & 0.84 & 0.88 & 0.82 & 0.89 & 0.84 
& 0.94 & 0.87 & 0.79 & 0.84 & 0.74 & 0.84 & 0.92 & 0.86 \\
Precision 
& \textit{1.00} & \textit{0.99} & 1.00 & 1.00 & 0.98 & 0.95 & 0.94 & 0.97 & 0.96 & 0.97 
& 0.98 & 0.99 & 0.77 & 0.76 & 0.71 & 0.75 & 0.97 & 0.93 \\
Recall    
& \textit{0.95} & \textit{0.90} & 0.40 & 0.40 & 0.79 & 0.72 & 0.76 & 0.66 & 0.79 & 0.70 
& 0.90 & 0.86 & 0.82 & 0.85 & 0.83 & 0.81 & 0.88 & 0.86 \\
F1-score  
& \textit{0.98} & \textit{0.94} & 0.57 & 0.59 & 0.88 & 0.82 & 0.86 & 0.78 & 0.87 & 0.81 
& 0.94 & 0.92 & 0.79 & 0.80 & 0.76 & 0.75 & 0.92 & 0.89 \\
\hline
\end{tabular}
\label{tab:main_results}
\end{table*}

Table~\ref{tab:main_results} compares the performance of all classification models under the aforementioned SF and MF settings. In SF, each timestamp contains at most one faulty variable, whereas in MF, multiple faults occur at the same timestamp simultaneously. Overall, most models achieve strong performance in SF, with MLP achieving the highest F1-score of 0.98 and TST achieving an F1-score of 0.94. Tree-based methods such as RF, XGBoost, and LightGBM also demonstrate competitive performance, with F1-scores ranging from 0.86 to 0.88.

\begin{table}[t]
\centering
\footnotesize
\setlength{\tabcolsep}{4pt}
\caption{Transformer models under multi-fault conditions.}
\renewcommand{\arraystretch}{1.15}
\begin{tabular}{c|c|cc|cc|cc}
\hline
\multirow2{*}{\textbf{Models}}
& \multirow2{*}{\textbf{LR}}
& \multicolumn{2}{c|}{\textbf{Precision}}
& \multicolumn{2}{c|}{\textbf{Recall}}
& \multicolumn{2}{c}{\textbf{F1-score}} \\
\cline{3-8}
&
& {{\textbf{W=16}}}& {{\textbf{W=32}}}
& {{\textbf{W=16}}}& {{\textbf{W=32}}}
& {{\textbf{W=16}}} & {{\textbf{W=32}}} \\
\hline
\multirow{3}{*}{{TST}}
& 1e-3 & 0.98 & 0.98 & 0.85 & 0.86 & \textit{0.91} & \textit{0.92} \\
& 3e-3 & 0.97 & 0.96 & 0.85 & 0.86 & \textit{0.9}1 & \textit{0.91} \\
& 5e-4 & 0.99 & 0.99 & 0.84 & 0.85 & \textit{0.91} & \textit{0.91} \\
\hline
\multirow{3}{*}{{PatchTST}}
& 1e-3 & 0.75 & 0.76 & 0.83 & 0.83 & 0.79 & 0.79 \\
& 3e-3 & 0.77 & 0.76 & 0.83 & 0.84 & 0.80 & 0.80 \\
& 5e-4 & 0.75 & 0.75 & 0.81 & 0.83 & 0.78 & 0.79 \\
\hline
\multirow{3}{*}{{Timer-S1}}
& 1e-3 & 0.61 & 0.70 & 0.78 & 0.80 & 0.69 & 0.75 \\
& 3e-3 & 0.70 & 0.75 & 0.70 & 0.74 & 0.70 & 0.74 \\
& 5e-4 & 0.59 & 0.59 & 0.81 & 0.81 & 0.68 & 0.68 \\
\hline
\multirow{3}{*}{{TimesNet}}
& 1e-3 & 0.90 & 0.92 & 0.86 & 0.81 & 0.88 & 0.86 \\
& 3e-3 & 0.93 & 0.88 & 0.84 & 0.85 & 0.89 & 0.87 \\
& 5e-4 & 0.91 & 0.91 & 0.83 & 0.83 & 0.86 & 0.87 \\
\hline
\end{tabular}
\label{tab:compact_results}
\end{table}

Moreover, to explore the potential benefits of temporal modeling, we evaluated the performance of transformer models under different parameter settings in MF conditions, summarized in Table~\ref{tab:compact_results}. Among the models, TST consistently achieved the strongest performance, obtaining F1-scores up to 0.92 across different configurations. TimesNet also demonstrated stable behavior with F1-scores between 0.85 and 0.89, while PatchTST and Timer-S1 exhibited comparatively lower performance under certain settings. MF conditions introduce additional complexity, although their effect varies across models. Several models maintain strong performance, including MLP (F1 = 0.94), TST (F1 = 0.92), and TimesNet (F1 = 0.89). These results indicate that the anomaly classifier remains capable of producing reliable anomaly labels even when multiple faults occur concurrently.

Overall, the results demonstrate that the DT-driven anomaly detection pipeline can accurately identify both SF and MF conditions within the CPS testbed. Since the objective of \textit{AgenticTwin} is interpretable anomaly reasoning rather than detection itself, the best-performing classifier, MLP, is subsequently used to generate the structured anomaly labels consumed by the LLM-powered agentic interface of the framework.

\subsection{Benchmark Query Generation}

To systematically evaluate \textit{AgenticTwin}'s reasoning capabilities, we construct an LLM-assisted benchmark containing a total of 12,000 synthetic operator queries generated from anomaly events. We chose a powerful, state-of-the-art proprietary reasoning model, Gemini 2.5 Pro~\cite{comanici2025gemini} for this task. Each event is represented using its observed sensor values, DT predictions, residual features, anomaly labels, and fault metadata. These structured events are provided to an LLM through task-specific prompt templates that instruct it to generate realistic operator questions and corresponding reference responses. We generated 4,000 queries for each of the three task-specific subordinate agents (DA, RA, and MA respectively) in \textit{AgenticTwin}:

\begin{itemize}
    \item Diagnosis queries: Seeking explanations of anomaly causes and fault characteristics.
    \item Retrieval queries: Asking whether similar anomalies have previously occurred.
    \item Mitigation queries: Requesting corrective actions or mitigation recommendations.
\end{itemize}

For each query, Gemini 2.5 Pro also generates task-specific reference responses, using the injected anomaly metadata, DT outputs, and curated anomaly knowledge base. The generated queries and reference responses are subsequently reviewed for consistency and factual alignment, and the invalid samples are removed before evaluation.

\subsection{Evaluation of the Agentic LLM Interface}

We evaluate 15 LLMs in total, including the open-source Qwen~\cite{yang2025qwen3}, Llama~\cite{grattafiori2024llama}, Gemma~\cite{team2024gemma}, Phi~\cite{abdin2024phi}, DeepSeek R1 Distill~\cite{guo2025deepseek}, spanning small (\(\leq\)3B), medium (4--8B), and large (\(>\)8B) size groups. We also include one stronger proprietary baseline in our experiment, GPT 5.5~\cite{singh2025openai}.

\begin{table}[ht]
\centering
\caption{Semantic similarity scores of the DA and MA without and with the knowledge base and their relative gain.}
\label{tab:merged}
\fontsize{8.5}{9.5}\selectfont
\setlength{\tabcolsep}{2pt}
\renewcommand{\arraystretch}{1}
\begin{tabular}{@{}llcccccc@{}}
\toprule
\textbf{Size Group} & \textbf{Model}
& \multicolumn{3}{c}{\textbf{Diagnosis}}
& \multicolumn{3}{c}{\textbf{Mitigation}} \\
\cmidrule(lr){3-5}
\cmidrule(lr){6-8}
& & \textbf{WKB} & \textbf{KB} & \textbf{G(\%)}
  & \textbf{WKB} & \textbf{KB} & \textbf{G(\%)} \\
\midrule
\multirow{5}{*}{Small}
& Qwen 2.5 0.5B It      & 0.48 & 0.64 & 33.3
                     & 0.42 & 0.60 & 42.9 \\
& Llama 3.2 1B It        & 0.59 & 0.76 & 28.8
                     & 0.50 & 0.69 & 38.0 \\
& Gemma 3 1B It          & 0.53 & 0.71 & 34.0
                     & 0.46 & 0.66 & 43.5 \\
& Qwen 2.5 3B It         & 0.67 & 0.84 & 25.4
                     & 0.61 & 0.80 & 31.1 \\
& Llama 3.2 3B It        & 0.61 & 0.79 & 29.5
                     & 0.57 & 0.77 & 35.1 \\
\midrule
\multirow{5}{*}{Medium}
& Qwen 3 4B           & 0.69 & 0.85 & 23.2
                     & 0.66 & 0.84 & 27.3 \\
& Phi 3.5 Mini It 4B     & 0.63 & 0.82 & 30.2
                     & 0.59 & 0.80 & 35.6 \\
& Qwen 2.5 7B It         & 0.74 & 0.88 & 18.9
                     & 0.71 & 0.87 & 22.5 \\
& Qwen 3 8B           & 0.71 & 0.87 & 22.5
                     & 0.69 & 0.86 & 24.6 \\
& DS-R1-Llama-8B     & 0.77 & 0.90 & 16.9
                     & 0.73 & 0.88 & 20.5 \\
\midrule
\multirow{4}{*}{Large}
& DS-R1-Qwen-14B     & 0.80 & 0.91 & 13.8
                     & 0.77 & 0.90 & 16.9 \\
& Gemma 3 14B It         & 0.75 & 0.89 & 18.7
                     & 0.74 & 0.89 & 20.3 \\
& DS-R1-Qwen-32B     & 0.84 & 0.93 & 10.7
                     & 0.82 & 0.92 & 12.2 \\
& Llama 3.3 70B It      & 0.82 & 0.93 & 14.6
                     & 0.81 & 0.92 & 14.8 \\
\midrule
Proprietary
& GPT-5.5            & \textit{0.89} & \textit{0.96} & \textit{7.9}
                     & \textit{0.88} & \textit{0.96} & \textit{9.1} \\
\midrule
Average
& All models
& 0.69 & 0.84 & 22.3
& 0.66 & 0.84 & 26.3 \\
\bottomrule
\end{tabular}
\end{table}

Table~\ref{tab:merged} compares the semantic similarity scores 
achieved by the Diagnosis Agent (DA) and Mitigation Agent (MA) without (WKB) and with (KB) access to the anomaly knowledge base, along with the respective gains (G\%). Across all evaluated models, the inclusion of the knowledge base consistently improves reasoning quality. On average, the DA's scores increase from 0.69 to 0.84, corresponding to an average gain of 22.3\%, while the MA's scores increase from 0.66 to 0.84, yielding an average gain of 26.3\%. The gains are notable among smaller models, e.g., Qwen 2.5 0.5B Instruct achieves diagnosis and mitigation improvements of 33.3\% and 42.9\%, respectively, while Gemma 3 1B Instruct improves by 34.0\% and 43.5\%. Similar trends are observed for Phi 3.5 Mini Instruct 4B, which achieves gains exceeding 30\% on both tasks. These results indicate that lightweight models benefit substantially from access to structured anomaly-specific knowledge.

As model size increases, the relative gains generally decrease. Large models such as DeepSeek-R1-Distill-Qwen 32B and Llama 3.3 70B Instruct already exhibit strong performance without external grounding and therefore receive smaller relative improvements. GPT-5.5 achieves the highest absolute performance, reaching semantic similarity scores of 0.96 for both diagnosis and mitigation tasks. However, its relative gains are the smallest among all evaluated models, suggesting that much of the required knowledge for the evaluated tasks is already encoded within the model parameters. More importantly, several open-source models substantially narrow the performance gap after KB integration. For instance, DeepSeek-R1-Distill-Qwen 32B reaches diagnosis and mitigation scores of 0.93 and 0.92, respectively, approaching the performance of GPT-5.5 despite being considerably smaller and open-source. These findings demonstrate that providing access to the curated anomaly knowledge base significantly improves diagnosis and mitigation quality. Furthermore, results show that lightweight open-source models can achieve strong performance when provided with structured domain grounding.

\begin{table}[ht]
\centering
\caption{Performance evaluation of the RA.}
\label{tab:retrieval}
\fontsize{8.5}{9.5}\selectfont
\setlength{\tabcolsep}{2.5pt}
\renewcommand{\arraystretch}{1.0}
\begin{tabular}{@{}llccccc@{}}
\toprule
\textbf{Size Group} & \textbf{Model} & \textbf{P@1} & \textbf{R@1} & \textbf{P@3} & \textbf{R@3} & \textbf{MRR} \\
\midrule
\multirow{5}{*}{Small}
& Qwen 2.5 0.5B It   & 0.63 & 0.28 & 0.58 & 0.51 & 0.68 \\
& Llama 3.2 1B It   & 0.71 & 0.32 & 0.64 & 0.57 & 0.74 \\
& Gemma 3 1B It      & 0.67 & 0.30 & 0.61 & 0.55 & 0.71 \\
& Qwen 2.5 3B It    & 0.78 & 0.39 & 0.71 & 0.65 & 0.81 \\
& Llama 3.2 3B It   & 0.74 & 0.36 & 0.68 & 0.62 & 0.78 \\
\midrule
\multirow{5}{*}{Medium}
& Qwen 3 4B It      & 0.82 & 0.43 & 0.75 & 0.70 & 0.85 \\
& Phi 3.5 Mini It 4B & 0.79 & 0.41 & 0.73 & 0.68 & 0.83 \\
& Qwen 2.5 7B It    & 0.86 & 0.49 & 0.80 & 0.76 & 0.89 \\
& Qwen 3 8B       & 0.83 & 0.46 & 0.77 & 0.73 & 0.86 \\
& DS-R1-Llama-8B & 0.88 & 0.51 & 0.82 & 0.78 & 0.91 \\
\midrule
\multirow{4}{*}{Large}
& DS-R1-Qwen-14B & 0.90 & 0.54 & 0.84 & 0.81 & 0.93 \\
& Gemma 3 14B It     & 0.87 & 0.52 & 0.81 & 0.78 & 0.90 \\
& DS-R1-Qwen-32B & 0.92 & 0.58 & 0.86 & 0.84 & 0.95 \\
& Llama 3.3 70B It   & 0.91 & 0.56 & 0.85 & 0.82 & 0.93 \\
\midrule
Proprietary
& GPT-5.5        & \textit{0.94} & \textit{0.61}
                 & \textit{0.89} & \textit{0.87}
                 & \textit{0.97} \\
\midrule
Average & All models & 0.83 & 0.45 & 0.76 & 0.71 & 0.85 \\
\bottomrule
\end{tabular}
\end{table}

Table~\ref{tab:retrieval} reports that across all models, the average scores reach 0.83 P@1, 0.76 P@3, and an MRR of 0.85, indicating that the repository consistently returns relevant historical anomaly cases. The retrieval results exhibit a smaller performance gap across model scales than those for diagnosis and mitigation tasks. While GPT-5.5 achieves the strongest overall performance (MRR = 0.97), several open-source models achieve comparable retrieval quality. For example, Llama 3.3 70B Instruct achieves 0.93 and even medium-sized models such as Qwen 3 8B achieve an MRR of 0.86. The strong retrieval performance provides contextual evidence that can subsequently be utilized by the MA.

\subsection{Agentic versus Monolithic Reasoning}

This study examines whether decomposing anomaly analysis into specialized agents improves performance relative to performing all reasoning tasks with a single monolithic LLM. We compare \textit{AgenticTwin} with four monolithic baselines spanning different model scales: Qwen 2.5 3B Instruct, Qwen 2.5 7B Instruct, DeepSeek-R1-Distill-Llama 8B, and Llama 3.3 70B Instruct. These models have been chosen based on their task-specific performances in Tables~\ref{tab:merged} and~\ref{tab:retrieval} while covering multiple model scales.

For each monolithic baseline, a single LLM receives the complete information available for an anomaly event, including the observed and DT-predicted values, residual features, anomaly-classifier output, anomaly knowledge base, and historical anomaly repository. Through one unified prompt, the model is instructed to perform all three evaluated tasks: (i) anomaly diagnosis, (ii) retrieval of relevant historical cases, and (iii) mitigation recommendation. Its response is generated in a structured format so that the diagnosis, retrieval, and mitigation components can be evaluated separately using the corresponding metrics defined previously. 

In contrast, \textit{AgenticTwin} decomposes these tasks across three lightweight models for this experiment. Qwen 2.5 3B Instruct serves as the DA, Qwen 2.5 7B Instruct serves as the RA, and DeepSeek-R1-Distill-Llama 8B serves as the MA. Each agent receives only the information required for its assigned task, and the output of one stage is made available to subsequent stages when needed. Combined, the three models (i.e., \textit{AgenticTwin} in this experiment) contain 18B parameters.

\begin{table}[ht]
\centering
\caption{Monolithic baselines versus \textit{AgenticTwin}.}
\label{tab:mas_vs_single}
\fontsize{8.5}{9.5}\selectfont
\setlength{\tabcolsep}{1.25pt}
\renewcommand{\arraystretch}{1.1}
\begin{tabular}{@{}l|c|c|c|c|>{\centering\arraybackslash}p{1.75cm}@{}}
\toprule
&
\multicolumn{4}{c|}{\textbf{Monolithic Baselines}}
& \\
\cmidrule(lr){2-5}
\textbf{Metric}
&
\makecell{\textbf{Q 2.5}}
&
\makecell{\textbf{Q 2.5}}
&
\makecell{\textbf{DS-R1}}
&
\makecell{\textbf{L 3.3}}
&
{\textbf{\textit{AgenticTwin}}}
\\
\midrule
Model Size (B)
& 3 & 7 & 8 & 70 & 18$^\dagger$ \\
\midrule
Diagnosis (Similarity)
& 0.61 & 0.64 & 0.64 & 0.70 & \textit{0.93} \\
Retrieval (MRR)
& 0.70 & 0.68 & 0.71 & 0.74 & \textit{0.95} \\
Mitigation (Similarity)
& 0.69 & 0.73 & 0.77 & 0.78 & \textit{0.92} \\
\bottomrule
\end{tabular}
\vspace{1pt} \parbox{\columnwidth}{
{$^\dagger$\textit{AgenticTwin} comprises three specialized agents with 3B, 7B, and 8B parameters (total: 18B assigned parameters).}
}
\end{table}

Table \ref{tab:mas_vs_single} shows that \textit{AgenticTwin} achieves the highest performance across all three tasks, obtaining diagnosis similarity, retrieval MRR, and mitigation similarity scores of 0.93, 0.95, and 0.92, respectively. In comparison, the strongest monolithic baseline, Llama 3.3 70B Instruct, reaches 0.70, 0.74, and 0.78 on the same tasks. \textit{AgenticTwin} therefore improves over the best-performing baseline by 32.9\% in diagnosis, 28.4\% in retrieval, and 17.9\% in mitigation. These results indicate that the observed improvements are not attributable solely to model size. Instead, decomposing anomaly analysis into the specialized DA, RA, and MA allows the framework to process task-specific evidence more effectively. These findings demonstrate the benefit of agent specialization and show that coordinated lightweight open-source models can outperform a substantially larger standalone model. This indicates \textit{AgenticTwin}'s potential for resource-constrained CPS deployments.

\section{Conclusion}
\label{section: conclusion}

This paper presented \textit{AgenticTwin}, a digital twin-integrated agentic LLM framework for interpretable anomaly analysis in cyber-physical systems. The framework extends conventional anomaly detection with specialized agents for diagnosis, historical retrieval, and mitigation reasoning. A curated anomaly knowledge base and historical repository provide domain information and prior cases that support more contextual and actionable responses for system operators. We also introduced a benchmark constructed by injecting anomalies into a real-world weather monitoring dataset to support controlled evaluation of LLM-based anomaly reasoning. Experimental results show that access to the anomaly knowledge base improves diagnosis and mitigation performance across the evaluated LLMs, with the largest relative gains observed for lightweight, open-source models. Furthermore, the lightweight \textit{AgenticTwin} architecture outperforms all evaluated monolithic baselines, indicating that task specialization can potentially compensate for limited model scale. Overall, the findings demonstrate the potential of combining DT-based system modeling with lightweight LLMs for interpretable CPS decision support.

\section*{Acknowledgment} This work was supported by the National Science Foundation under Grant Number CNS-2423248.

\vspace{5px}
\noindent \textbf{Disclosure of AI use.} An AI-based language model (ChatGPT and Grammarly) was used to improve the grammar and readability of the manuscript. No scientific content was generated by the tool, and the authors retain full responsibility for the accuracy and integrity of the work.

\bibliographystyle{IEEEtran}
\bibliography{guin-bib,other-bib}

\vspace{-25px}

\begin{IEEEbiography}[{\includegraphics[width=1\linewidth]{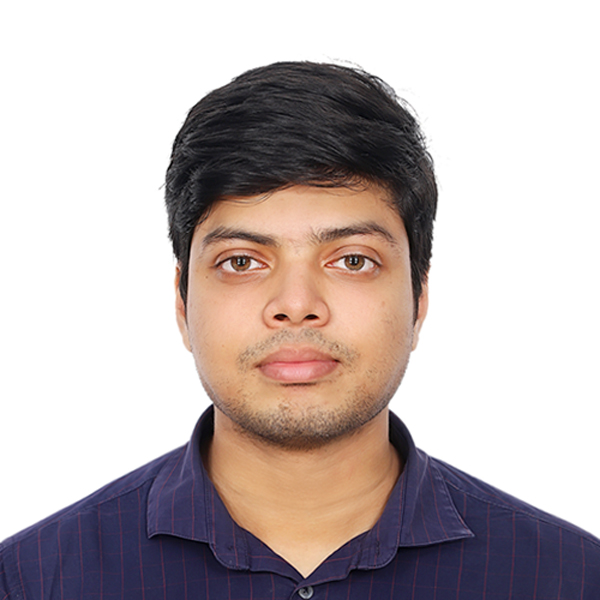}}]{Touseef Hasan (S'25)} is pursuing his Ph.D. from the Department of Electrical and Computer Engineering in the School of Computing at Wichita State University, Wichita, KS, USA. He received his B.Sc. from Bangladesh University of Engineering and Technology, Dhaka, Bangladesh in 2024. His research interests include exploring the practical applications of LLMs in hardware security and digital twins. He is a student member of the IEEE.
\end{IEEEbiography}

\vspace{-25px}

\begin{IEEEbiography}[{\includegraphics[width=0.9\linewidth]{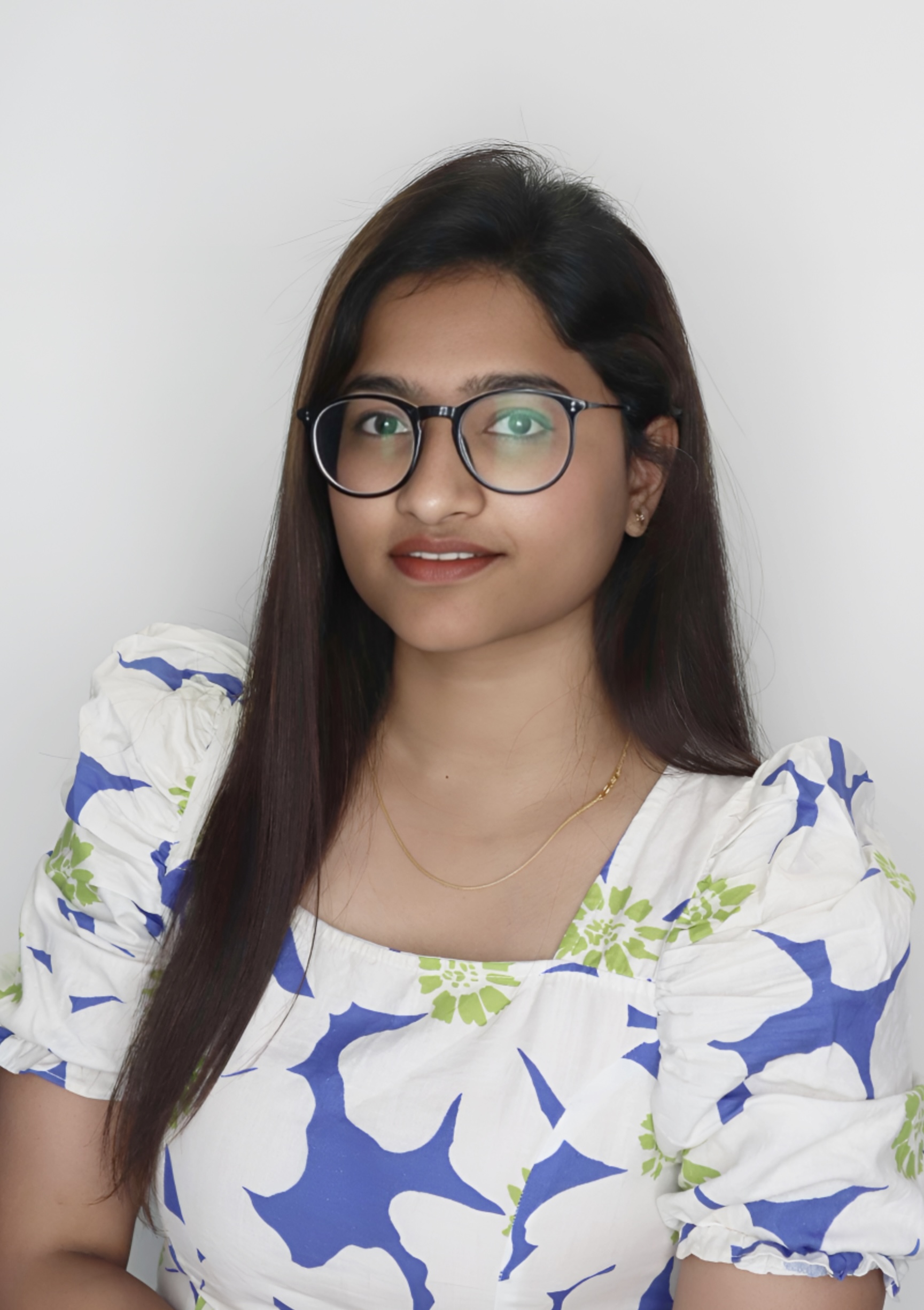}}]{Mounika Ghanta (S'26)} received her M.S in Computer Science and Software Engineering from Auburn University, AL, USA in 2025. She is currently pursuing her Ph.D. in Electrical Engineering from the Department of Electrical and Computer Engineering at Auburn University. She received the B.Tech. degree in Electronics and Communication Engineering from Hyderabad Institute of Technology and Management, India, in 2023. Her research interests include hardware security, digital twins, and LLMs. She is a student member of the IEEE.
\end{IEEEbiography}

\vspace{-25px}

\begin{IEEEbiography}[{\includegraphics[width=1in, clip, keepaspectratio]{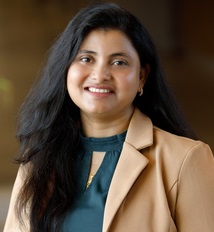}}]{Souvika Sarkar (M'25)} received her Ph.D. in Computer Science and Software Engineering from Auburn University in 2020. She is currently an Assistant Professor in the School of Computing at Wichita State University, Wichita, KS, USA. Prior to that, she completed her B.Tech in Computer Science and Engineering from West Bengal University of Technology, India, in 2012, and earned her M.E. in Software Engineering from Jadavpur University in 2018. 

Dr. Sarkar’s research interests include natural language processing (NLP), generative AI, LLMs, and building scalable and adaptive AI systems. Her research centers on developing efficient NLP systems, with a strong focus on interpretability, adaptability, and real-world impact. She has published widely in top-tier venues in AI and NLP. She actively serves the research community as a reviewer for leading conferences, including ACL, EMNLP, NAACL, and AACL.
\end{IEEEbiography}

\vspace{-25px}

\begin{IEEEbiography}[{\includegraphics[width=1\linewidth, clip, keepaspectratio]{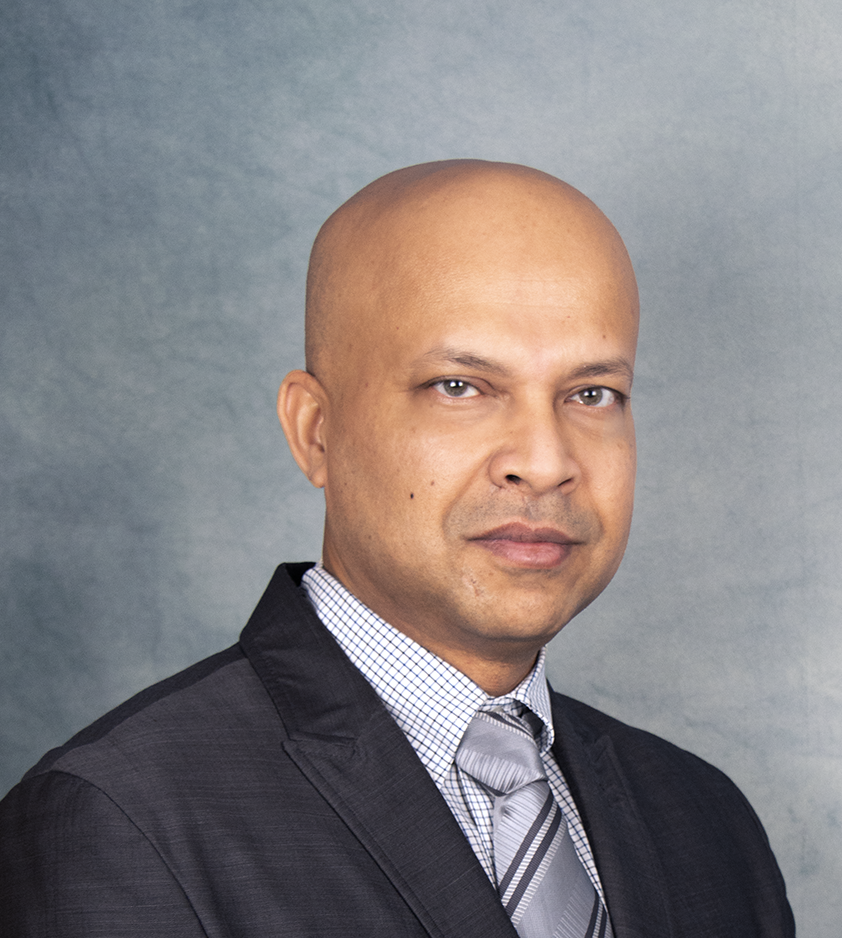}}]{Ujjwal Guin (S'10--M'16--SM'22)} received his Ph.D. in Electrical and Computer Engineering from the University of Connecticut in 2016. He is currently the Godbold Associate Professor in the Department of Electrical and Computer Engineering at Auburn University, Auburn, AL, USA. He earned his B.E. in Electronics and Telecommunication Engineering from Bengal Engineering and Science University, India, in 2004, and his M.S. in Electrical and Computer Engineering from Temple University, Philadelphia, PA, USA, in 2010.

Dr. Guin's research interests span hardware security and trust, supply chain security, cybersecurity, and VLSI design and test. He has developed a broad range of on-chip structures and techniques to enhance the security, trustworthiness, and reliability of integrated circuits. He has authored numerous journal articles and refereed conference papers, several of which have earned best paper nominations and awards. His research has been funded by the National Science Foundation (NSF) and multiple Department of Defense (DoD) institutions. He actively contributes to the academic community by serving on organizing and technical program committees for leading conferences, including HOST, ITC, VTS, and DAC.

\end{IEEEbiography}

\end{document}